\documentclass[letterpaper]{article}
\usepackage{aaai}
\usepackage{times}
\usepackage{helvet}
\usepackage{courier}
\usepackage{natbib}

\usepackage[utf8]{inputenc} 
\usepackage[T1]{fontenc}    
\usepackage{hyperref}       
\usepackage{url}            
\usepackage{booktabs}       
\usepackage{amsfonts}       
\usepackage{nicefrac}       
\usepackage{microtype}      

\usepackage{graphicx} 
\usepackage{caption}
\usepackage{subcaption}
\graphicspath{{figures/}} 

\usepackage{csquotes} 

\usepackage{amsmath}
\usepackage{flexisym} 

\DeclareMathOperator*{\argmax}{arg\,max}
\usepackage{bm} 

\usepackage{tikz}
\newcommand*\circled[1]{\tikz[baseline=(char.base)]{
    \node[shape=circle,draw,inner sep=0.6pt] (char) {#1};}}

\begin{document}
%
\title{$k$-Nearest Neighbor Augmented Neural Networks for Text Classification}
\author{
Zhiguo Wang\textsuperscript{1}, Wael Hamza\textsuperscript{1}, Linfeng Song\textsuperscript{2}\\
\textsuperscript{1} IBM T.J. Watson Research Center, Yorktown Heights, NY 10598 \\
\textsuperscript{2} Department of Computer Science, University of Rochester, Rochester, NY 14627\\
}
\maketitle
\begin{abstract}
\begin{quote}
In recent years, many deep-learning based models are proposed for text classification.
This kind of models well fits the training set from the statistical point of view. 
However, it lacks the capacity of utilizing instance-level information from individual instances in the training set. 
In this work, we propose to enhance neural network models by allowing them to leverage information from $k$-nearest neighbor (kNN) of the input text. 
Our model employs a neural network that encodes texts into text embeddings.
Moreover, we also utilize $k$-nearest neighbor of the input text as an \textit{external memory}, and utilize it to capture instance-level information from the training set. 
The final prediction is made based on features from both the neural network encoder and the kNN memory. 
Experimental results on several standard benchmark datasets show that our model outperforms the baseline model on all the datasets, and it even beats a very deep neural network model (with 29 layers) in several datasets. 
Our model also shows superior performance when training instances are scarce, and when the training set is severely unbalanced.
Our model also leverages techniques such as  
semi-supervised training and transfer learning quite well.
\end{quote}
\end{abstract}

\section{Introduction}
\label{sec:introduction}
Text classification is a fundamental task in both natural language processing and machine learning research. 
Its goal is to assign specific labels to texts written in natural languages. Based on the definition of the specific labels, text classification has many practical applications, 
e.g., sentiment analysis \citep{xia2013dual} and news categorization \citep{li2009framework}.

In recent years, with the renaissance of neural networks, many deep-learning based methods were proposed for text classification tasks \citep{zhang2015character,conneau2016very,joulin2016bag,johnson2016supervised}.
Basically, most of these methods construct some kinds of neural networks to encode a text into a distributed text embedding, and then predict the category of the text solely based on it. 
In the training stage, network parameters are optimized on the training set. 
In the testing stage, the entire training set can be discarded, and only the trained model is used for prediction. 
This method has acquired the state-of-the-art performance in many tasks. 
However, because it abstracts the training set from a statistical point of view, it cannot utilize instance-level information from individual instances in the training set very well. 
For example, in the news categorization task, the following news is annotated as the ``Business'' category.
\begin{displayquote}
\textit{Eastman Kodak Company and IBM will work together to develop and manufacture image sensors used in such consumer products as digital still cameras and camera phones .}
\end{displayquote}
A neural network model with the state-of-the-art performance will incorrectly predict it into the ``Sci/Tech'' category, 
because, in the training set,  up to 1,166 instances about ``IBM'' are annotated as the ``Sci/Tech'' category, 
whereas only 278 instances about ``IBM'' are annotated as the ``Business'' category. 
From the statistical point of view, we cannot blame our model, because the single word ``IBM'' is a very strong signal for the ``'Sci/Tech' category. 
However, when we look at the 278 instances with the ``Business'' category, we found a much relevant instance:
\begin{displayquote}
\textit{IBM and Eastman Kodak Company have agreed to jointly develop and manufacture image sensors for mass - market consumer products , such as digital still cameras .}
\end{displayquote}
Therefore, if we can make use of category information of the relevant training instance, we will have a big chance to correct the error.

On the other hand, instance-based (or non-parametric) learning \citep{aggarwal2014instance} provides us a good method to capture instance-level information. 
$k$-nearest neighbor (kNN) classification is the most representative method, where a predication is made for a new test instance only based on its kNN. 
\citet{quinlan1993combining} showed that a better performance can be achieved if combining the model-based learning and the instance-based learning.

Therefore, in this work, we propose to enhance neural network models with information from kNN. 
Our model still employs a neural network encoder to abstract global information from the entire training set, and to encode texts into text embeddings. 
Moreover, we also take the kNN of the input text as an \textit{external memory}, and utilize it to capture instance-level information from the training set. 
Then, the final prediction is made based on features from both the neural network encoder and the kNN memory. 
Concretely, for each input text, we first find its kNN in the training set. 
Second, a neural network encoder is utilized to encode both the input text and its kNN into text embeddings. 
Third, based on the text embeddings of the input text and the kNN, we calculate attention weights for each neighbor. 
Based on these attention weights, the model calculates an \textit{attentive kNN label distribution} and an \textit{attentive kNN text embedding}. 
The final prediction is made based on three sources of features: the text embedding of the input text, the \textit{attentive kNN label distribution}, and the \textit{attentive kNN text embedding}. 
Experimental results on several standard benchmark datasets show that our model outperforms the baseline model on all the datasets, and it even beats a very deep neural network model (with 29 layers) in several datasets. 
Our model also shows superior performance when training instances are scarce, and when the training set is severely unbalanced.
Our model also leverages techniques such as  
semi-supervised training and transfer learning quite well.

In following parts, we start with the description of our model, then evaluate the model on some standard benchmark datasets and different experimental settings. We then talk about related work, and finally conclude this work.

\begin{figure*}[t]
  \centering
  \includegraphics[width=1.0\linewidth]{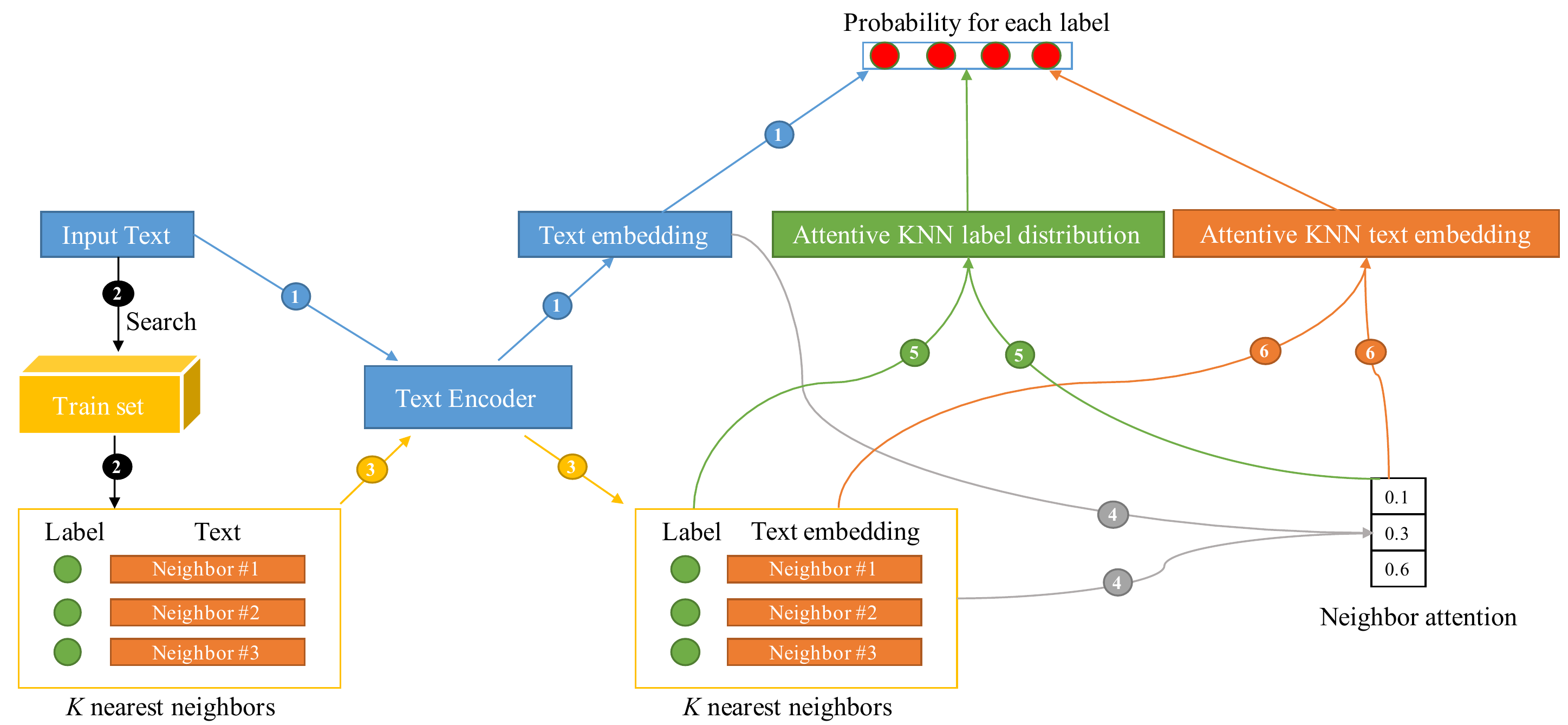}
  \caption{Overall architecture of our model.}
  \label{fig:architecture}
\end{figure*}

\section{Model}
\label{sec:model}

In this section, we propose a model to capture both global and instance-level information from the training set for text classification tasks. 
To capture global information, we train a neural network encoder to encode texts into an embedding space based on all training instances and their category information. 
To capture instance-level information, for each input text, we search its kNN from the training set, and then take the kNN as an external memory to augment the neural network.

Figure \ref{fig:architecture} shows the architecture of our model. 
The blue flow {\color{blue}\circled{1}} is the typical method for text classification,  where an input text is encoded into a text embedding by a neural network ``Text Encoder'', and then a prediction is made based on the text embedding. 
The remaining flows are our kNN memory, which employs the attention mechanism to extract some instance-level features for prediction. Formally, given an input text $x$, its kNN \{$x\textprime_1$,...,$x\textprime_k$,...,$x\textprime_K$\} and their correct labels $y$ and \{$y\textprime_1$,...,$y\textprime_k$,...,$y\textprime_K$\}, our task can be formulated as estimating a conditional probability $\Pr{(y|x,x\textprime_1,...,x\textprime_K,y\textprime_1,...,y\textprime_K)}$ based on the training set, and predicting the labels for testing instances by 
\begin{equation}
y^* = \argmax_{y \in \mathcal{A}(y)} \Pr(y|x,x\textprime_1,...,x\textprime_K,y\textprime_1,...,y\textprime_K), 
\label{eq:preds}
\end{equation}
where $\mathcal{A}(y)$ is a set of all possible labels.

\subsection{Text Encoder}
\label{subsec:encoder}
Text Encoder is a critical component in both typical models and our model. Its task is to encode an input text into a text embedding. Typically, an encoder encodes a text in two steps: (1) \textit{word representation step} represents all words in the text as word embeddings or character embeddings \citep{zhang2015character}; and (2) \textit{sentence representation step} composes the word embedding sequence into a fixed-length text embedding with the Convolutional Neural Networks (CNN) \citep{lecun1998gradient} or the Long Short-Term Memory Network (LSTM) \citep{hochreiter1997long} models. For example, \citet{kim2014convolutional}, \citet{kalchbrenner2014convolutional}, and \citet{wang2016sentence} employed the CNN model to encode texts, \citet{wang2016semi} utilized the LSTM model to represent texts, and \citet{lai2015recurrent} combined both the CNN and the LSTM.

In this work, we utilize a LSTM network to encode texts. For \textit{word representation step}, 
inspired by \citet{wang2016multi} and \citet{wang2017bilateral}, 
we construct a $d$-dimensional vector for each word with two components: a word embedding and a character-composed embedding. The word embedding is a fixed vector for each individual word, which is pre-trained with GloVe \citep{pennington2014glove} or word2vec \citep{mikolov2013distributed}. The character-composed embedding is calculated by feeding each character (also represented as a vector) within a word into a LSTM. For \textit{sentence representation step}, we apply a bi-directional LSTM (BiLSTM) to compose the word representation sequence, and then concatenate the two vectors from the last time-step of the BiLSTM (both the forward and the backward directions) as the final text embedding.

\subsection{kNN Memory}
\label{subsec:external}
kNN memory is the core component of our model. 
Its goal is to capture instance-level information for each input text from its kNN. This component includes the following six procedures.

\paragraph{Searching for the kNN} 
This procedure, corresponding to the black flow \circled{2} in Figure \ref{fig:architecture}, is to find the kNN of the input text from the training set.
In order to efficiently search over the large training set, we employ a traditional information retrieve method to find the kNN. 
We first build an \textit{inverted index} for all texts in the training set with the open source toolkit Lucene (https://lucene.apache.org/). 
Then, we take the input text as the query, and utilize the simple and effective BM25 ranking function \citep{robertson2009probabilistic} to retrieve the kNN from the \textit{inverted index}.

\paragraph{Encoding for the kNN}
This procedure encodes all the kNN into text embeddings, which corresponds to the yellow flow {\color{yellow}\circled{3}} in Figure \ref{fig:architecture}. We re-utilize the Text Encoder described above, and apply it to each of the $K$ neighbors individually.

\paragraph{Calculating the neighbor attention}
This procedure corresponds to the gray flow {\color{gray}\circled{4}} in Figure \ref{fig:architecture}, and its goal is to calculate similarities (\textit{neighbor attention}) between the input text and each of the $K$ neighbors in the embedding space.
Formally, let's denote the text embeddings for the input text $x$ and the $k$-th neighbor $x\textprime_k$ as $\bm{h}$ and $\bm{h}\textprime_k$, which are $l$-dimensional vectors calculated by the Text Encoder.
Theoretically, any similarity metrics will fit here.
Inspired from \cite{wang2016multi, wang2017bilateral},
here, we adopt the effective \textit{multi-perspective cosine matching} function  $f_s$ to compute similarities between two vectors $\bm{h}$ and $\bm{h}\textprime_k$: 
\begin{equation}
\bm{s}_k = f_{s}(\bm{h},\bm{h}\textprime_k;\bm{W})
\label{equ:MP_cosine}
\end{equation}
where $\bm{W} \in \Re^{I \times l}$ is a trainable parameter with the shape $I \times l$, $I$ is a hyper-parameter to control the number of perspectives, and the returned value $\bm{s}_k$ is a $I$-dimensional vector $\bm{s}_k=[s_k^1,...,s_k^i,...,s_k^I]$. Each element $s_k^i \in \bm{s}$ is a similarity between $\bm{h}$ and $\bm{h}\textprime_k$ from the $i$-th perspective, and it is calculated by the cosine similarity between two weighted vectors
\begin{equation}
s_k^i=cosine(W_i \circ \bm{h}, W_i \circ \bm{h}\textprime_k)
\label{equ:weight_cosine}
\end{equation}
where $\circ$ is the element-wise multiplication, and $W_i$ is the $i$-th row of $\bm{W}$, which controls the $i$-th perspective and assigns different weights to different dimensions of the $l$-dimensional text embedding space.

We set $I=1$ for the illustration in Figure \ref{fig:architecture}, therefore the neighbor attention is just a vector and each neighbor has only one similarity to the input text. However, for the experiments in following sections, we will utilize multiple perspectives ($I>1$), and each neighbor could have multiple similarities to the input text.

\paragraph{Calculating the attentive kNN label distribution} Based on the neighbor attentions, we calculate the \textit{attentive kNN label distribution} by weighted summing up the label distributions of all kNN (the green flow {\color{green}\circled{5}} in Figure \ref{fig:architecture}).
Formally, let's denote the label distribution of the $k$-th neighbor $x\textprime_k$ as $\bm{y}\textprime_k$, which is an one-hot $c$-dimensional vector for the correct label $y\textprime_i$, and $c$ is the number of all possible labels in the classification task. Given the label distributions and the neighbor attentions of all kNN, we calculate the $i$-th perspective of the \textit{attentive kNN label distribution} by
\begin{equation}
\bm{\hat{y}}_i=\sum_{k=1}^{K} s_k^i * \bm{y}\textprime_k
\label{equ:att-label-distribution}
\end{equation}
where $*$ is an operation to multiply the left scalar with each element of the right vector.
Then, the final \textit{attentive kNN label distribution} $\bm{\hat{y}}$ is the concatenation of \{$\bm{\hat{y}}_1$,...,$\bm{\hat{y}}_i$,...,$\bm{\hat{y}}_I$\} from all $I$ perspectives.

\paragraph{Calculating the attentive kNN text embedding}
Similarly, the \textit{attentive kNN text embedding} is the weighted sum of text embeddings of all kNN (the orange flow {\color{orange}\circled{6}} in Figure \ref{fig:architecture}).
Given the text embeddings and neighbor attentions of all kNN, we calculate the $i$-th perspective of the \textit{attentive kNN text embedding} by
\begin{equation}
\bm{\hat{h}}_i=\sum_{k=1}^{K} s_k^i * \bm{h}\textprime_k
\label{equ:att-text-embedding}
\end{equation}
Then, the final \textit{attentive kNN text embedding} $\bm{\hat{h}}$ is the concatenation of \{$\bm{\hat{h}}_1$,...,$\bm{\hat{h}}_i$,...,$\bm{\hat{h}}_I$\} from all $I$ perspectives.

\paragraph{Concatenating all features to make the prediction}
As the final procedure, we concatenate three sources of features (or vectors): the input text embedding $\bm{h}$, the \textit{attentive kNN label distribution} $\bm{\hat{y}}$, and the \textit{attentive kNN text embedding} $\bm{\hat{h}}$. Then, a fully-connected layer with the \textit{softmax} function is applied to make the final prediction.

\section{Experiments}
\label{sec:experiments}

\paragraph{Datasets}

\begin{table*}[t]
  \centering
  \begin{tabular}{lcccc}
    \toprule
    Dataset & \# Classes & Train Samples & Dev Samples & Test Samples \\
    \midrule
    AG's News & 4 & 118,000 & 2,000 & 7,600\\
Sogou News & 5 & 447,500 & 2,500 & 60,000\\
DBPedia & 14 & 553,000 & 7,000 & 70,000\\
Yelp Review Polarity & 2 & 559,000 & 1,000 & 38,000\\
Yelp Review Full & 5 & 647,500 & 2,500 & 50,000\\
Yahoo!Answers & 10 & 1,395,000 & 5,000 & 60,000\\
Amazon Review Full & 5 & 2,997,500 & 2,500 & 650,000\\
Amazon Review Polarity & 2 & 3,599,000 & 1,000 & 400,000\\
    \bottomrule
  \end{tabular}
  \caption{Statistics of the datasets.}
  \label{tab:datasets}
\end{table*}

We evaluate our model on eight publicly available datasets from \citet{zhang2015character}. Here are the brief descriptions for all datasets.
\begin{itemize}
\item AG's News: This is a news categorization dataset. All news articles are obtained from the AG's corpus of news article on the web. Each news belongs to one out of the four labels \{\textit{World}, \textit{Sports}, \textit{Business}, \textit{Sci/Tech}\}.
\item Sogou News: This is a Chinese news categorization dataset. All news articles are collected from SogouCA and SogouCS news corpora \citep{wang2008automatic}. Each news article belongs to one out of the five categories \{\textit{sports}, \textit{finance}, \textit{entertainment}, \textit{automobile}, \textit{technology}\}. All Chinese characters have been transformed into \textit{pinyin} (a phonetic romanization of Chinese).
\item DBPedia: This dataset is designed for classifying Wikipedia articles into 14 ontology classes from DBpedia. Each instance contains the title and the abstract of a Wikipedia article.
\item Yelp Review Polarity/Full: This is a sentiment analysis dataset. All reviews are obtained from the Yelp Dataset Challenge in 2015. Two classification tasks are constructed from this dataset. The first one predicts the number of stars the user has given, and the second one predicts a polarity label by considering stars 1 and 2 as negative, and 3 and 4 as positive.
\item Yahoo! Answers dataset: This is a topic classification dataset. The dataset is obtained from Yahoo! Answer Comprehensive Question and Answer version 1.0 dataset. Each instance includes the question title, the question content and the best answer. And each instance belongs to one out of 10 topics.
\item Amazon Reviews Polarity/Full: This is another sentiment analysis dataset. All reviews are obtained from the Stanford Network Analysis Project (SNAP). Similar to the Yelp Review dataset, a Full version and a Polarity version of the dataset are constructed.
\end{itemize}
The original datasets didn't provide the devsets. To avoid tuning model parameters on the test sets, for each dataset, we build a devset by randomly holding out 500 instances for each class from the original training set, and take the remaining instances as our new training set. Table \ref{tab:datasets} shows the statistics of all the datasets.

\paragraph{Experiment Settings}
We initialize word embeddings with the 300-dimensional GloVe word vectors pre-trained from the 840B Common Crawl corpus \citep{pennington2014glove}. 
For the out-of-vocabulary (OOV) words, we initialize their word embeddings randomly.  
For the character-composed embeddings, we represent each character with a 20-dimensional randomly-initialized vector, and feed characters of each word into a LSTM layer to produce a 50-dimensional vector. 
We set the hidden size to 100 for our BiLSTM Text Encoder.
We train the entire model from end-to-end, and minimize the cross entropy of the training set. We use the ADAM optimizer \citep{kingma2014adam} to update parameters, and set the learning rate as 0.0001. 
During training, we do not update the pre-trained word embeddings. For all experiments, we iterate over the training set for 15 times, and evaluate the model on the devset at the end of each iteration. Then, we pick the model which works the best on the devset as the \textit{final} model, and all the results on the test sets are performed from the \textit{final} models.

\begin{table*}[t]
  \centering
  \begin{tabular}{clc}
    \toprule
    Model ID & Feature Configuration &	Accuracy\\
    \midrule
M1 & text-embedding & 91.9 \\
M2 & attentive-kNN-label & 91.5\\
M3 & attentive-kNN-text & 92.2 \\
M4 & attentive-kNN-label + attentive-kNN-text & 93.2 \\
M5 & text-embedding + attentive-kNN-label & 93.8\\
M6 & text-embedding + attentive-kNN-text & 93.6\\
M7 & text-embedding + attentive-kNN-label + attentive-kNN-text & 94.6\\
    \bottomrule
  \end{tabular}
  \caption{Evaluation of different feature configurations.}
  \label{tab:feature_config}
\end{table*}

\begin{figure}[th]
\centering
\includegraphics[width=1.0\linewidth]{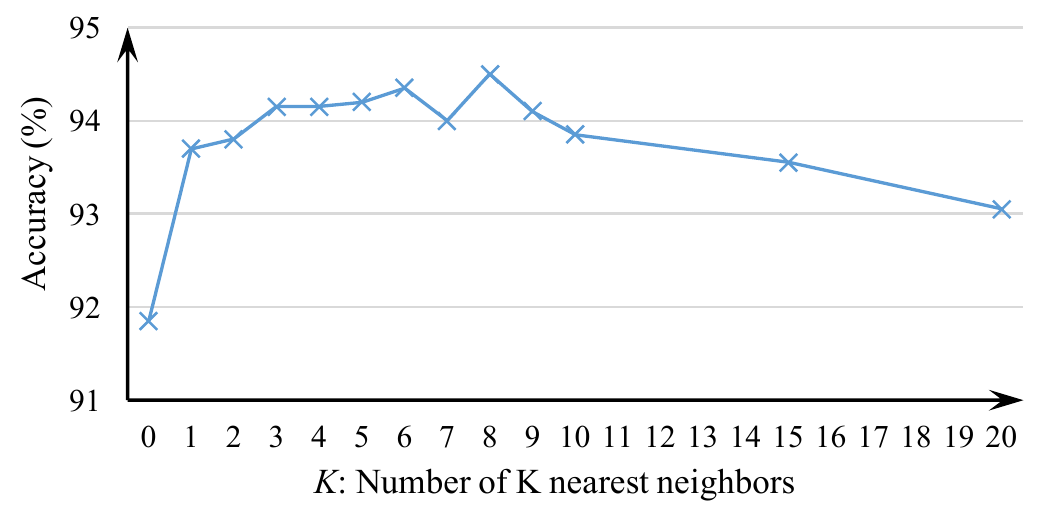}
\captionof{figure}{Influence of K nearest neighbors.}
\label{fig:knn}
\end{figure}

\begin{figure}[th]
\centering
\includegraphics[width=1.0\linewidth]{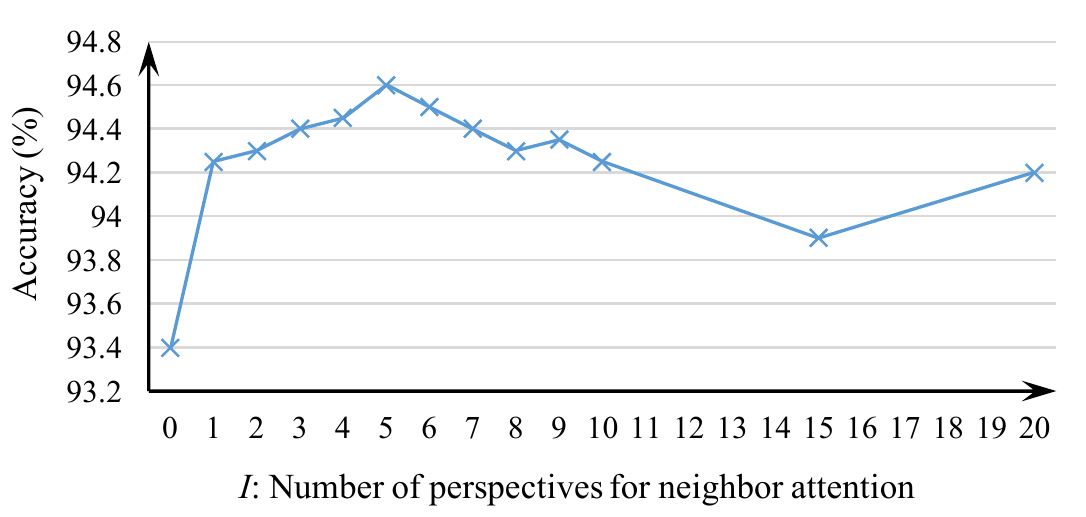}
\captionof{figure}{Influence of multi-perspective attentions.}
\label{fig:mp-cosine}
\end{figure}

\subsection{Properties of our Model}
There are several hyper-parameters in our model.
The choices of them may affect the final performance. In this subsection, we conduct some experiments to demonstrate the properties of our model, and select a group of proper hyper-parameters for subsequent experiments.
All experiments in this subsection are conducted on the AG's News dataset, and evaluated on the devset.

First, we study the effectiveness of the three sources of features: the input text embedding (text-embedding), the attentive kNN label distribution (attentive-kNN-label), and the attentive kNN text embedding (attentive-kNN-text). 
We create seven models according to different feature configurations. 
For all models, we set $K=20$ for the number of kNN, and $I=20$ for the number of perspectives in our multi-perspective cosine matching function (Equation (\ref{equ:MP_cosine})).  
Table \ref{tab:feature_config} shows the corresponding results on the devset, where M1 is our BiLSTM model without using the kNN memory, M2/M3/M4 are the models only utilizing the kNN memory, and M5/M6/M7 are the models leveraging both the text encoder and the kNN memory.
From this experiment, we get several interesting observations: 
(1) when comparing M2 with M1, we find that only utilizing the label information from the kNN can achieve a competitive performance to the typical BiLSTM model;
(2) the performance from M3 is on par with M1, which indicates that features solely extracted from the kNN (but without label information) can represent the input text very well;
(3) the performance of M4 shows that considering both the label and the text information from the kNN achieves a better performance than the typical BiLSTM model, which shows the effectiveness of our kNN memory;
(4) from the results of M5/M6/M7, we can see that combining the kNN memory with our BiLSTM text encoder achieves even better accuracies.
The best accuracy is obtained by M7. 
Therefore, we will use this configuration for the subsequent experiments.

Second, to test the influence of the kNN, we vary $K$ from 1 to 20. 
Figure \ref{fig:knn} shows the accuracy curve, where $K=0$ corresponds to the performance from our BiLSTM without using the kNN memory. 
We can see that even with only one neighbor ($K=1$), our model gets a significant improvement over the BiLSTM model.
When increasing the number of neighbors, the performance improves at the beginning, and then drops when $K$ exceeds 8. 
One possible reason is that the neighbors become noisy when increasing $K$. In the subsequent experiments, we fix $K=5$.

Third, we investigate the influence of the hyper-parameter $I$ in our multi-perspective cosine matching function (Equation (\ref{equ:MP_cosine})) by varying $I$ from 1 to 20. 
We build a baseline by replacing Equation (\ref{equ:MP_cosine}) with the vanilla cosine similarity function. Both of the baseline and our model with $I=1$ calculate a single attention for each neighbor, but the difference is that our model assigns some trainable parameters to each dimension of the embedding space.
Figure \ref{fig:mp-cosine} shows the accuracy curve, where $I=0$ corresponds to the performance of our baseline.
We find that even if utilizing only one perspective ($I=1$), our model achieves a significant improvement than the baseline. 
When increasing the number of perspectives, the accuracy improves at the beginning, and then decreases after $I$ is over 5. 
Therefore, we fix $I=5$ in the subsequent experiments.

\subsection{Comparison with the State-of-the-art Models}
\begin{table*}[t]
  \centering
  \begin{tabular}{lcccccccc}
    \toprule
    Model & AG & Sogou & DBP & Yelp P. & Yelp F. & Yah. A & Amz. F. & Amz. P.\\
    \midrule
 BoW$^a$ & 88.8 & 92.9 & 96.6 & 92.2 & 58.0 & 68.9 & 54.6 & 90.4\\
ngrams$^a$ & 92.0 & 97.1 & 98.6 & 95.6 & 56.3 & 68.5 & 54.3 & 92.0\\
ngrams-TFIDF$^a$ & 92.4 & \textbf{97.2} & 98.7 & 95.4 & 54.8 & 68.5 & 52.4 & 91.5\\
 char-CNN$^a$ & 87.2 & 95.1 & 98.3 & 94.7 & 62.0 & 71.2 & 59.5 & 94.5\\
char-CRNN$^b$ & 91.4 & 95.2 & 98.6 & 94.5 & 61.8 & 71.7 & 59.2 & 94.1\\
VDCNN$^c$ & 91.3 & 96.8 & 97.7 & \textbf{95.7} & \textbf{64.7} & 73.4 & \textbf{63.0} & \textbf{95.7}\\
fastText-unigram$^d$ & 91.5 & 93.9 & 98.1 & 93.8 & 60.4 & 72.0 & 55.8 & 91.2\\
fastText-bigram$^d$ & 92.5 & 96.8 & 98.6 & \textbf{95.7} & 63.9 & 72.3 & 60.2 & 94.6\\ 
    \midrule
BiLSTM & 92.5 & 94.4 & 98.9 & 92.4 & 59.3 & 72.5 & 59.0 &94.7 \\
BiLSTM with kNN & \textbf{94.2} & 96.5& \textbf{99.1} & 94.5 & 61.9 & \textbf{74.4} & 60.3& 95.3\\
    \bottomrule
  \end{tabular}
  \caption{Evaluation on the test sets, and the state-of-the-art models by \citet{zhang2015character}$^a$, \citet{xiao2016efficient}$^b$, \citet{conneau2016very}$^c$, and \citet{joulin2016bag}$^d$.}
  \label{tab:accuracy}
\end{table*}

We construct two models to evaluate on all of the test sets. 
The first model is the baseline: our BiLSTM model without using the kNN memory (M1 in Table \ref{tab:feature_config}). 
The second model is the BiLSTM model with the kNN memory (M7 in Table \ref{tab:feature_config}). 
Table \ref{tab:accuracy} gives the experimental results. 
We find that by utilizing the kNN memory, our \textit{BiLSTM with kNN} model outperforms the baseline on all datasets. 
Among all the state-of-the-art models, the VDCNN model \citep{conneau2016very} is a very deep network with up to 29 convolutional layers.
Our model even beats the VDCNN model on the AG's News, DBpeida and Yahoo! Answers datasets, which shows the effectiveness of our method. Moreover, our kNN memory can be easily adapted into these more complex neural network models.

\subsection{Evaluation in Other Training Setups}
To study the behaviors of our model, we further evaluate it in some other common training setups. 
All the experiments in this subsection are conducted on the AG's News dataset, and the accuracies are performed on the devset.

\begin{table*}[t]
  \centering
  \begin{tabular}{lccc}
    \toprule
Model & Full Setup & Low-Resource Setup & Unbalanced Setup\\
    \midrule
BiLSTM & 91.9 & 85.2 (-6.7) & 48.6 (-43.3) \\
BiLSTM with kNN & 94.6 & 90.2 (-4.4) & 90.6 (-4.0)\\
    \bottomrule
  \end{tabular}
  \caption{Evaluation in the rare-resource setup and the unbalanced training setup, where the numbers in brackets show the change of accuracy against the ``Full Setup''.}
  \label{tab:unbalanced}
\end{table*}

\paragraph{Low-Resource Training Setup}
In the introduction section, we claimed that the kNN memory captures instance-level information from the training set. 
To verify this claim, we evaluate our model on a low-resource training setup. 
We construct a low-resource training set by randomly selecting 10\% of all instances for each category from the original training set.
Then, we train our ``BiLSTM'' and ``BiLSTM with kNN'' models, with the same configurations as before, on this low-resource training set. 
In Table \ref{tab:unbalanced}, the third column, with the title ``Low_Resource Setup'',  shows the accuracies of our two models. Comparing with the models trained with the full training set (``Full Setup''), our ``BiLSTM with kNN'' model only drops 4.4 percent, which is lower than the 6.7 percent drop from our ``BiLSTM'' model. 
This result shows our kNN memory can capture instance-level information to remedy shortage of the low-resource training set.

\paragraph{Unbalanced Training Setup} 
In this sub-section, we evaluate our model on an unbalanced training set, a scenario not uncommon in text classification. 
To severely skew the label distribution, we construct an unbalanced training set by randomly selecting 2,000 instances for the \textit{World} category, 4,000 instances for the \textit{Sports} category, 8,000 instances for the \textit{Business} category and 16,000 instances for the \textit{Sci/Tech}. 
We train both our BiLSTM and proposed models (with the same configuration as before) on this unbalanced training set. 
The last column of Table \ref{tab:unbalanced} shows the accuracies. 
The BiLSTM model shows a severe (43.3\%) degradation of accuracy in the unbalanced setting.
On the other hand, our \textit{BiLSTM with kNN} model only drops 4.0\% when trained on the unbalanced training set. 
We believe this is because our model can capture instance-level information from the unbalanced training set.

\begin{table}[t]
  \centering
  \begin{tabular}{lccc}
    \toprule
    Setup & Accuracy \\
    \midrule
BiLSTM (M1) & 91.9 \\
Semi-supervised Training (M6) & 92.9\\
Transfer Learning (M5) & 92.7\\
Transfer Learning (M7) & 93.4\\
    \bottomrule
  \end{tabular}
  \caption{Evaluation in the semi-supervised training and the transfer learning setups.}
  \label{tab:semi-supervised}
\end{table}

\paragraph{Semi-supervised Training and Transfer Learning} 
So far, we have verified that the effectiveness of the kNN retrieved from the same training set.
\textit{What if we search the kNN from a dataset of a completely different task}? 
If we consider the text (without the label) information of the kNN from a different task, then this becomes the \textbf{semi-supervised training} setup. 
If we utilize the label information of the kNN from a different task, where the definition of labels are quite different from the task at hand, then this becomes the \textbf{transfer learning} setup. 
To reveal the behavior of our model in these two setups, we take the training set of the DBPedia dataset as the corpus to find the kNN for each text in the AG's News dataset. 
We construct four models: ``BiLSTM'' using the ``M1'' configuration in Table \ref{tab:feature_config},
``Semi-supervised Training'' using the ``M6'' configuration, ``Transfer Learning (M5)'' using the ``M5'' configuration, 
and ``Transfer Learning (M7)'' using the ``M7'' configuration.
Here, we should notice that the ``attentive-kNN-text'' and the ``attentive-kNN-label'' features are extracted from the kNN from the DBPedia corpus, which is a different classification task.
Table \ref{tab:semi-supervised} gives the results.
We can see that our model achieves improvements from the baseline (BiLSTM) in both the \textit{Semi-supervised Training} and the \textit{Transfer Learning} setups.

\subsection{Qualitative Analysis}
We perform qualitative analysis by looking at some instances incorrectly predicted by the baseline (\textit{BiLSTM}) but get corrected by adding the kNN memory (\textit{BiLSTM with kNN}).

First, for the error illustrated in the introduction section, our model corrected it as expected.

Another example is ``\textit{President George W. Bush 's campaign website was inaccessible from outside the United States .}'' with the correct label \textit{Sci/Tech}. 
In the training set, the appearances of the phrase ``George W. Bush'' in each category are 1,659 (World), 410 (Business), 67 (Sports) and 299 (Sci/Tech), and the frequencies of the word ``President'' in each category are 4,486 (World), 1,021 (Business), 233 (Sports) and 554 (Sci/Tech). 
Based on these strong signals, the BiLSTM baseline assigned it with an incorrect label \textit{World}. 
On the other hand, our \textit{BiLSTM with kNN} corrected it, because there is a very similar neighbor in the training set ``\textit{President George W. Bush 's official re-election website was down and inaccessible for hours , in what campaign officials said could be the work of hackers .}'' with the label \textit{Sci/Tech}.

\section{Related Work}
\label{sec:related}
In recent years, there have been several studies trying to augment neural network models with external memories. Generally, these models utilize the attention mechanism to access useful information outside the model itself (so-called the external memory). 
For the machine translation task, \citet{bahdanau2014neural} introduced the attention mechanism to access source side encoding information while generating the target sequence. 
For the question answering task, \citet{weston2014memory} proposed the memory network to access all supporting sentences before generating the correct answer word. 
\citet{graves2014neural} designed a more complicated ``computer-like'' external memory to simulate the Turning Machine. 
Our model also belongs to this group, but with the distinction that we construct the external memory with the $K$ nearest neighbors of the input text, and utilize a multi-perspective attention model.

\citet{vinyals2016matching} proposed a matching network for one shot learning task. 
Their model also classifies the input instance by utilizing a labeled support set as our model does. 
However, our model differentiates from their model in two ways. 
First,  they assume the labeled support set is given beforehand, whereas our model searches the kNN independently based on the input instance. 
Second, their model only utilizes the label information of the support set for prediction, whereas our model makes use of information from both the input text and the kNN.

Our model follows an old idea of combining the model-based (or parametric) learning and the instance-based (or non-parametric) learning \citep{quinlan1993combining}. We infuse the old idea with the advanced neural networks and the attention mechanism.

\section{Conclusion}
 In this work, we enhanced neural networks with a kNN memory for text classification. Our model employs a neural network encoder to abstract information from the entire training set, and utilizes the kNN memory to capture instance-level information. 
The final prediction is made based on features from both the input text and the kNN.
Experimental results on several standard benchmark datasets show that our model outperforms the baseline model on all the datasets, and it even beats a very deep neural network model (with 29 layers) in several datasets. 
Our model also shows superior performance when training instances are scarce, and when the training data is severely unbalanced. Our model also leverages techniques such as 
semi-supervised training and transfer learning quite well.

\bibliography{aaai}
\bibliographystyle{aaai}

\end{document}